\documentclass{article}
\usepackage{spconf,amsmath,epsfig}

\begin{document}\sloppy

\def\x{{\mathbf x}}
\def\L{{\cal L}}

\title{Unsupervised Semantic Deep Hashing}
%
\name{Sheng Jin}
\address{}

\maketitle

\begin{abstract}
In recent years, deep hashing methods have been proved to be efficient since it employs convolutional neural network to learn features and hashing codes simultaneously. However, these methods are mostly supervised. In real-world application, it is a time-consuming and overloaded task for annotating a large number of images. In this paper, we propose a novel unsupervised deep hashing method for large-scale image retrieval. Our method, namely unsupervised semantic deep hashing (\textbf{USDH}), uses semantic information preserved in the CNN feature layer to guide the training of network. We enforce four criteria on hashing codes learning based on VGG-19 model: 1) preserving relevant information of feature space in hashing space; 2) minimizing quantization loss between binary-like codes and hashing codes; 3) improving the usage of each bit in hashing codes by using maximum information entropy, and 4) invariant to image rotation. Extensive experiments on CIFAR-10, NUSWIDE have demonstrated that \textbf{USDH} outperforms several state-of-the-art unsupervised hashing methods for image retrieval. We also conduct experiments on Oxford 17 datasets for fine-grained classification to verify its efficiency for other computer vision tasks.
\end{abstract}
\section{Introduction}
With the explosive increase of data, searching for content relevant image, or other media data remains a challenge because of large amount of computational cost and the accuracy requirement. In the early stage, researchers focus on data-independent methods. Locality-Sensitive Hashing \cite{andoni2006Near} and its variants are proposed. But it has a lower accuracy since the semantic information of data is not considered during coding process. In recent years, the data-dependent hashing methods \cite{lin2014fast} attract more attention since its compact representation and superior accuracy performance. Compared with data-independent hashing method, data-dependent hashing methods improve retrieval performance via training on the dataset.

Data-dependent methods mainly include supervised hashing methods \cite{liu2012supervised}, unsupervised hashing methods \cite{weiss2009spectral} and semi-supervised hashing methods \cite{wang2010semi}. These supervised methods make use of the class information provided in the manual labels, where the supervised information is used in three forms: point-wise labels, pair-wise labels, and ranking labels. Some representative works have been proposed, e.g. Semantic Hashing \cite{salakhutdinov2009semantic}, Binary Reconstruction Embedding \cite{kulis2009learning}, Minimal Loss Hashing\cite{norouzi2011minimal}, Kernel-based Supervised Hashing \cite{liu2012supervised}, Hamming Distance Metric Learning \cite{norouzi2012hamming}, and Column Generation Hashing \cite{li2013learning}. Although the supervised hashing methods and semi-supervised hashing methods have been proved to gain better accuracy with compacter hashing codes, it is a time-consuming and heavy workload task in practical application. In the past years, some classical unsupervised hashing methods also have been developed, e.g. Isotropic Hashing \cite{kong2012isotropic}, Spherical Hashing \cite{heo2012spherical}, Discrete Graph Hashing \cite{liu2014discrete}, Locally Linear Hashing \cite{irie2014locally}, Asymmetric Inner-product Binary Coding \cite{shen2015learning} and Scalable Graph Hashing \cite{jiang2015scalable}.

In these traditional hashing methods, each image is initially represented by a hand-crafted feature. However, these features may not preserve accurate semantic information. And they also may not be suitable for generating binary codes. Due to these facts, the accuracy of image retrieval could not meet our requirement. Over the last five years, deep learning has been proved to be effective in computer vision because it could automatically extract high-level semantic feature to represent image that is robust to the variances of object. Hinton \cite{salakhutdinov2009semantic} et al. firstly proposed hashing method based on deep neural network. However, in \cite{salakhutdinov2009semantic}, the input of the network is still hand-crafted features, which is the most crucial limitation.

Very recently, convolutional Neural Network Hashing \cite{xia2014supervised} introduces an end-to-end network for learning better hashing codes. However, this method has limitations since it cannot perform feature learning and hashing codes learning simultaneously. Followed \cite{xia2014supervised}, new variants of deep hashing have been proposed, e.g, Deep Neural Network Hashing \cite{lai2015simultaneous}, Deep Semantic Ranking Hashing \cite{zhao2015deep}, deep supervised hashing \cite{liu2016deep} and DeepBit \cite{lin2016learning}, which extract features and learn hashing codes simultaneously. These methods are more effective and perform more efficiently in image retrieval task. However, most of these deep hashing methods, except DeepBit \cite{lin2016learning} and DBD-MQ \cite{Duan2017learning}, are pure supervised. DBD-MQ \cite{Duan2017learning} propose a quantization method for hashing learning. This method does not utilize the rigid sign function for binarization and considers the binarization as a multi-quantization task. DeepBit \cite{lin2016learning} tries to make hashing codes invariant to rotation by minimizing the difference between the hashing codes that describe the reference image and that of rotated one. However, this method only considers rotation invariance of images, and the invariance among different images with same class label can not be guaranteed.

In this paper, motivated by the success of DeepBit \cite{lin2016learning}, we propose a novel unsupervised deep hashing method, called unsupervised semantic deep hashing (\textbf{USDH}).
\begin{figure*}
  \centering
\includegraphics[height=1.75in, width=5.38in]{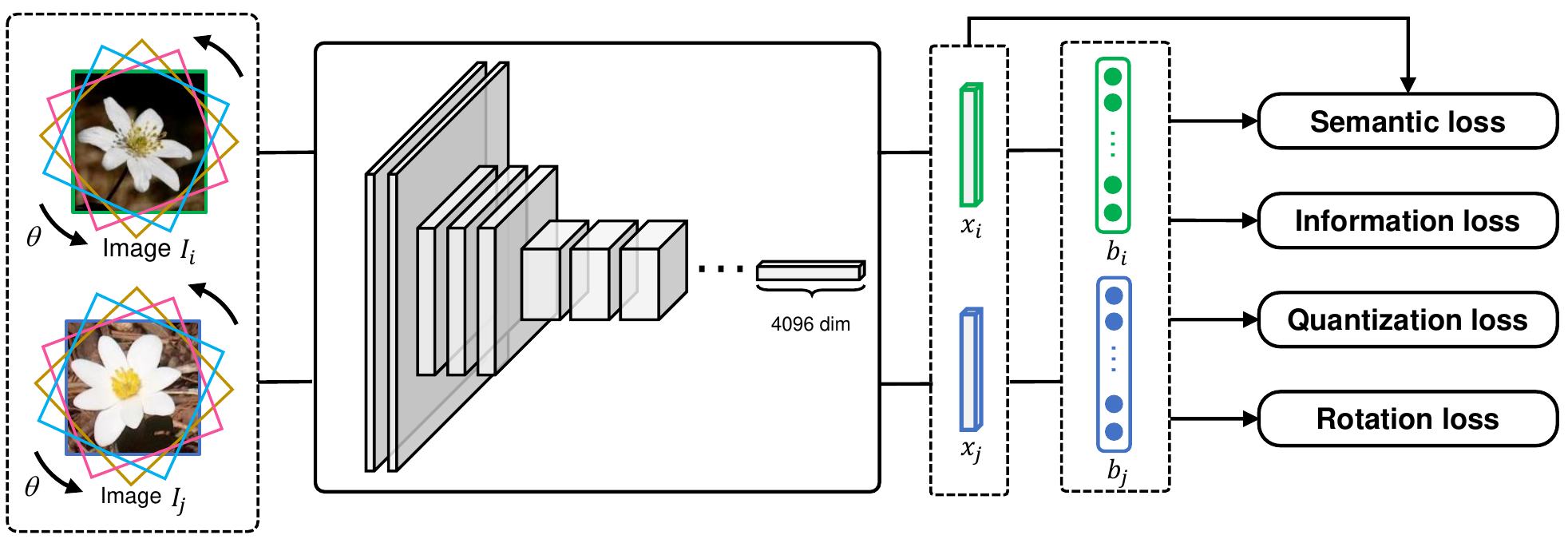}
\caption{We enforce four criterions on the loss function to learn efficient hashing codes based on VGG-19 architecture. In the training stage, the hashing codes are learned by the form of image pairs. On the first stage, we train the deep model by minimizing quantization loss ,information loss and use the mid-level feature to guide the process of  learning hashing codes. On the second stage,  we augment dataset with rotation ,  hashing codes is learned to be invariant to rotation by minimization the distance between that represents reference image and that of rotated one.}
\label{Fig2:}
\end{figure*}

The main contributions of \textbf{USDH} are outlined as follows:

\textbf{USDH} is an unsupervised end-to-end deep hashing framework. Compared with DeepBit method, \textbf{USDH} not only considers rotation invariance in a single image, but also preserves the semantic information of image pairs.

\textbf{USDH} proposes a novel deep unsupervised hashing method to preserve the similarity information in the feature space. It regards the output of full-connected layer as representation descriptor of image. The loss function requires hashing codes learned by deep network approximating the similarity computed by representation descriptors of image.

Experiments on general datasets show that \textbf{USDH} can outperform other unsupervised methods to achieve the state-of-the-art performance in image retrieval applications. And it is also quite effective for fine-grained classification.

\section{Unsupervised Semantic Deep Hashing}
In this paper, we introduce a novel unsupervised deep hashing method. Compared with existing methods, our method utilizes relevant information preserved in the feature space to guide the learning process of hashing codes. Based on this motivation, the cost function of  unsupervised semantic deep hashing contains four components: 1) preserving relevant information of feature space in hashing codes via a semantic loss 2) minimizing quantization loss between binary-like codes and hashing codes 3) improving the usage of each bit in hashing codes by maximizing information entropy 4) keeping the learned hashing codes invariant to rotation by pulling hashing codes of reference image and that of the rotated one together. The whole deep model is shown in Figure~\ref{Fig2:}. The cost function is written as below:

\begin{equation}
  J=J_{1}+J_{2}+J_{3}+J_{4},
\end{equation}
$J_{1}$ represents semantic loss, $J_{2}$ represents quantization loss, $J_{3}$ represents information loss, $J_{4}$ represents rotation loss.


\subsection{Semantic Loss}
To preserve semantic information in the feature space, firstly, we should adopt an optimal feature to represent images and use a proper formula to measure the similarity of images in the feature space, then we let similarity computed by the hashing codes of image pairs approximate the similarity measured in the original feature space.

Firstly, we adopt VGG-19 model to process the images and use the output of the second full-connected layer as our image feature. Many researches have proved that high-level feature of convolutional neural network has sufficient semantic information and these mid-features are robust to inner-class including rotation, shape and color variance. There also exist different metrics to measure similarity in the feature space. We adopt a widely-used metric that is defined as:

\begin{equation}
   S_{i,j}=e^{\frac {-{\Vert{x_{i}-x_{j}\Vert}}_{2}}{\rho d}},
\end{equation}
Where $d$ denotes the dimension of the second full-connected layer and $\rho$ is a positive constant parameter. $S_{i,j}\in(0,1]$ can represent a similarity degree of the images $i$ and $j$. The hashing codes of image $i$ is denoted as $b_i$.

We require hashing codes preserving relevant semantic information. More specifically, if $S_{i,j}$ is near to 1, we assume hashing codes $b_{i}$ and $b_{j}$ has smaller distance. But if $S_{i,j}$ is near to 0, then $b_{i}$ and $b_{j}$ has larger distance. For each training batch, we can obtain a similarity matrix. We try to use the similarity degree in the feature space to guide the learning of hashing codes. To do so, in hamming space, we also define a similarity measure, and then the similarity measure defined in hashing space is
required to be as similar as possible to the similarity matrix defined in the original feature space.

According to this constraint, the neighbor points in the feature space are still neighbors in the hashing space. Specifically, $b_{i}\in\{0,1\}$ is relaxed to $(0,1)$, then the hashing codes is linearly transformed to $(-1,1)$:
\begin{equation}
   \widetilde b_{i}=2b_{i}-1,
\end{equation}
where $\widetilde b_{i}\in(-1,1)$. The inner product of $\widetilde b_{i}$and $\widetilde b_{j}$ is in the range of $(-k,k)$, where $k$ is the length of hashing codes. Then the
inner product is linear transformed to $(0,1)$ via $\frac{\widetilde b_{i}^T*\widetilde b_{j}+k}{2k}$. The result of linear transformation is also regarded as a similarity degree. And it fits in with the assumption on information loss that each bit of hashing codes plays the same role.The function of semantic loss is written as:

\begin{equation}
    J_{1}=\sum_{i,j} {{\left \vert S_{i,j}-\frac{\widetilde b_{i}^T*\widetilde b_{j}+k}{2k}\right \vert}}_{1}.
\end{equation}
With this loss function, the deep model is trained by back-propagation algorithm with batch gradient descent method. To solve this, the gradient of semantic loss function need to be computed. Since $l_{1}$ norm is non-differentiable at some certain point , we employ sub-gradient to overcome the problem and we define the sub-gradient at this point to be equal to right-hand derivative. The gradient of semantic loss is defined as:

\begin{equation}
     \frac{\partial J_{1}}{\partial b_{i}} =\sum_{j} sgn(S_{i,j}-\frac{\widetilde b_{i}^T*\widetilde b_{j}+k}{2k})*{\frac{\widetilde b_{j}}{2k}}
\end{equation}
where

\[sgn(x)=\left\{\begin{array}{ll}
1& x \geq 0  ,\\
-1& x<0  .
\end{array}\right.\]

\subsection{Quantization Loss}
Since it is difficult to directly optimize discrete loss function, we should relax the objective function to transform the discrete problem into a continuous optimization problem.
As discussed in \cite{liu2016deep}, some widely-used relaxation scheme working with non-linear functions, such as sigmoid and tanh function,  would inevitably slow down or even restrain the convergence of the network \cite{krizhevsky2012imagenet}. To overcome such limitation, we still use the relu function as activation function of the second full-connected layer. Then the output of network is quantized to the binary codes. The quantization function is written as:

\[f(b_i)=\left\{\begin{array}{ll}
1& b_i \geq 0.5  ,\\
0& b_i<0.5  .
\end{array}\right.\]
where $f(x)$ denotes the binarization function.

To decrease this loss, we let the value of network's output near to 1 or 0. First, the hashing codes $b_i$ is linearly transformed to $(-1,1)$ in the same way. Then the result of linearly transformation is changed into an absolute value. The absolute value of the hashing codes $\vert \widetilde b_i \vert$ should be near to 1. Finally the quantization loss is defined as:

\begin{equation}
     J_2=\alpha\sum_{i} {\left \vert |\widetilde b_i|-1 \right \vert}_1,
\end{equation}
where$\left \vert . \right \vert $ denotes element-wise absolute value, and ${\left \Vert . \right \Vert}_1 $ denotes  $l_1$ norm. $\alpha$ is a weighting parameter.

To train the model, the gradient of  $J_2$  need to be computed. The sub-gradient is taken to replace the gradient of $J_2$ because of the non-differentiate point in the absolute operation and $l_1$ norm. The gradient is written as:

\[ \frac{\partial J_2}{\partial b_i}=\left\{\begin{array}{ll}
2\alpha       & b_i \geq 1\quad or\quad 0<b_i<0.5,\\
-2\alpha  & \quad otherwise  .
\end{array}\right.\]

\subsection{Information Loss}
As the main assumption of semantic loss, each bit of hashing codes should play an equivalent impact, which means each bit should have the same mean value. Inspired by the efficiency of DeepBit \cite{lin2016learning} method, we also maximize capability of each bit in hashing codes to express information. So we further enhance the hashing codes by assuming that each bit has half chance to be one. Based on this constraint, the balanced distribution criterion can be written as below:

\begin{equation}
     \mu_i=\frac{1}{m}\sum_{i=1}^{m}{b_i(m)},
\end{equation}
where $\mu_i$ denotes the mean value of $i$-th bite of hashing codes, ${\Vert . \Vert}_2 $ denotes $l_2$ norm and $m$ denotes the size of training batch.

\subsection{Rotation Loss}
Existing widely-used hand-crafted features should be invariant to rotation and scale. Inspired by this motivation, we also rotate the images and pull hashing codes that represent
the reference image and that of the rotated one together. The proposed rotation-invariance criterion can be written as:

\begin{equation}
     J_4=\sum_{i=1}^{m}\sum_{\theta=0}^{2R} {\left \Vert b_{\theta,i}-b_i\right \Vert},
\end{equation}
Where $b_{\theta,i}$ denotes hashing codes of image $i$ with rotation $\theta$.

\section{Experiment}
In order to test the performance of our proposed method, we conduct experiments on four datasets, including three widely used image retrieval datasets: CIFAR-10 and NUSWIDE dataset, as well as one recognition dataset: Oxford flower17. Similar to other image retrieval task, our method is also evaluated based on mean accuracy precision at top 1000. Compared with some representative unsupervised hashing methods, such as KMH \cite{He2013K}, SphH \cite{heo2012spherical}, SpeH \cite{weiss2009spectral}, PCAH \cite{wang2010semi}, LSH \cite{andoni2006Near}, PCA-ITQ \cite{gong2013iterative}, DH \cite{lin2015deep}, DeepBit \cite{lin2016learning} and DBD-MQ \cite{Duan2017learning}, experimental results verify that our proposed method outperforms these existing unsupervised hashing method. In order to prove our method is flexible for other computer vision applications, we also conduct experiments for fined-grained recognition on Oxford flower17 dataset.

\subsection{Dataset}
{\bfseries CIFAR-10 dataset} consists of 60000 32$\times$32 images in 10 classes. Each image in dataset belongs to one class.( 6000 images per class) The dataset is divided in two parts: train set(5000 images per class) and test set(1000 images per class).

\par\noindent {\bfseries NUSWIDE dataset} is a multi-label dataset. NUSWIDE contains nearly 270k images associated with 81 semantic concepts. Followed \cite{xia2014supervised}, We select the 21 most frequent concept. Each of concepts is associated with at least 5000 images. The dataset is splitted into training set and test set. We sample 100 images from each concepts to form a test set and the remaining images are treated as a training set.
\par\noindent {\bfseries Oxford 17 flower} dataset consists of 1360 images belonging to 17 mutually classes. Each class contains 80 images. The dataset is divided into three parts, including train set, test set and validation set, with 40 images, 20 images and 20 images respectively. In our experiment, we ignore validation set.

\subsection{Implementation Details}
The \textbf{USDH} method is implemented based on Caffe and the deep model is trained by batch gradient descend. As shown in Figure \ref{Fig2:}, We use VGG-19 as the base model, and the model is firstly trained on Imagenet dataset. Then the output layer of VGG-19 is replaced by hashing layer. In the training stage, image is regarded as input in the form of batch and every two images in same batch construct an image pair. The parameters of deep model are updated by minimizing objective function, including semantic loss, quantization loss, information loss and rotation loss. We conduct experiments for learning 16-bit, 32-bit, 48-bit hashing codes, respectively on cifar-10 dataset and NUSWIDE dataset. In this paper, we propose multiple loss function. So we further evaluate these loss functions. The semantic loss is proved more important and our quantization loss also improve performance. Since the efficiency of semantic loss, robustness analysis is discussed. We conduct experiments by different parameters $\rho$ in semantic loss. The constant parameters $\rho$ are respectively set as $d$, $\frac{d}{2}$, $\frac{d}{4}$. Where $d$ denotes as the dimension of output of second full-connected layer. To prove the efficiency of hashing codes learned by \textbf{USDH}, we also conduct experiments for other computer vision field, such as fined grained classification.
\subsection{Results on image retrieval}
Similar to DeepBit \cite{lin2016learning} method, the dataset is splitted into two parts. More specially, 10000 images is selected randomly as query image and then we conduct retrieval task on the remaining images for both CIFAR-10. We define similarity label based on semantic-level labels and images from the same class are considered similar. For NUSWIDE dataset, we follow the setting in \cite{xia2014supervised}, and if two images share at least one same label, they are considered same. The Mean Average Precision (MAP,\%)
at top 1000 of different unsupervised hashing methods on CIFAR-10 dataset was shown in table1. The experiment results on Table~\ref{tab1:} show that \textbf{USDH} outperforms existing best retrieval performance by $4.6\%$, $10.1\%$, $7.3\%$ and improves DeepBit method by $6.7\%$, $11.7\%$, $11.5\%$, correspond to different hash bits, respectively 16 bits, 32 bits and 64 bits. we also conduct experiments for large-scale image retrieval. As shown in Table~\ref{tab2:}, our method absolute increases of $25.77\%$, $25.58\%$, $24.67\%$ in average MAP for different bits on NUSWIDE dataset. Based on results of experiment, \textbf{USDH} is proved to be effective for image retrieval and the semantic information among different images in feature space improves significantly performance.

\begin{table}
  \centering
  \caption{Mean Average Precision (MAP) results for different number of bits CIFAR-10}
  \label{tab1:}
  \begin{tabular}{lccc}
  \hline
    Method&16-bit&32-bit&64-bit\\
  \hline
    KMH& 13.59& 13.93& 14.46\\
    SphH& 13.98& 14.58& 15.38\\
    SpeH& 12.55 & 12.42& 12.56\\
    PCAH& 12.91& 12.60& 12.10\\
    LSH& 12.55& 13.76& 15.07\\
    PCA-ITQ& 15.67& 16.20& 16.64\\
    DH& 16.17& 16.62& 16.96\\
    DeepBit & 19.43& 24.86& 27.73\\
    DBD-MQ & 21.53& 26.50& 31.85\\
  \hline
  {\bfseries USDH} & {\bfseries 26.13}& {\bfseries 36.56}& {\bfseries 39.27}\\
  \hline
  \end{tabular}
\end{table}
\begin{table}
  \centering
  \caption{MAP results for different number of bits NUSWIDE}
  \label{tab2:}
  \begin{tabular}{lccc}
    \hline
    Method&16-bit&32-bit&48-bit\\
    \hline
    SphH & 41.30& 42.40& 43.10\\
    SpeH & 43.30& 42.60& 42.30\\
    PCAH & 42.90& 43.70& 41.40\\
    LSH & 40.30& 42.60& 42.30\\
    PCA-ITQ & 45.28& 46.82& 47.70\\
    DH & 42.20& 44.80& 48.00\\
    DeepBit & 38.30& 40.10& 41.20\\
     \hline
  {\bfseries USDH}&{\bfseries 64.07}  & {\bfseries 65.68}  & {\bfseries 65.87}\\
   \hline
  \end{tabular}
\end{table}
{\bfseries Component analysis of loss function:} Our loss function consists of four components. In this section, we evaluate the effectiveness of two major components: semantic loss and quantization loss. The results on CIFAR-10 are shown in Table~\ref{tab6:} . It is worth mentioning that the semantic loss has improved the performance by 7.62\% compared to DeepBit method. And the quantization loss proposed in our paper has further improved the performance by 4.07\%.
\begin{table}
  \centering
  \caption{Effectiveness (MAP 32 bits) of different loss function}
  \label{tab6:}
  \begin{tabular}{lc}
    \hline
    Method& MAP\\
    \hline
    DeepBit& 24.86\\
    DeepBit+semantic loss& 32.48\\
    {\bfseries our method}& {\bfseries 36.55}\\
    \hline
  \end{tabular}
\end{table}

{\bfseries Robustness analysis of semantic loss:} Since all these experiment results have shown the effectiveness of semantic loss, the next experiment would focus on the influences of different parameter settings. We set the parameter $\rho$  in different value, including $1$, $\frac{1}{2}$, $\frac{1}{4}$, $\frac{1}{8}$, to conduct experiments on CIFAR10 to learn the 64-bits hashing codes, where $d$ denotes the dimension of second full-connected layer. Table~\ref{tab3:} reveals that semantic loss is robust to the value of parameter $\rho$. The experimental results suggest that the hashing codes learned by \textbf{USDH} focus on relative relationship of image features, instead of their exact similarity value.
\begin{table}
  \centering
  \caption{Comparison of image retrieval MAP of our \textbf{USDH} with respect to different
  values of parameters $\rho$}
  \label{tab3:}
  \begin{tabular}{ccccc}
    \hline
    $\rho$& $1$ & $1/2$ & $1/4$ & $1/8$\\
     \hline
     MAP& 39.27 & 39.02 & 39.20 & 39.11 \\
    \hline
  \end{tabular}
\end{table}


\subsection{Results on fined grained classification}
\begin{figure}
\centering
\includegraphics[height=0.62\linewidth,width=0.9\linewidth]{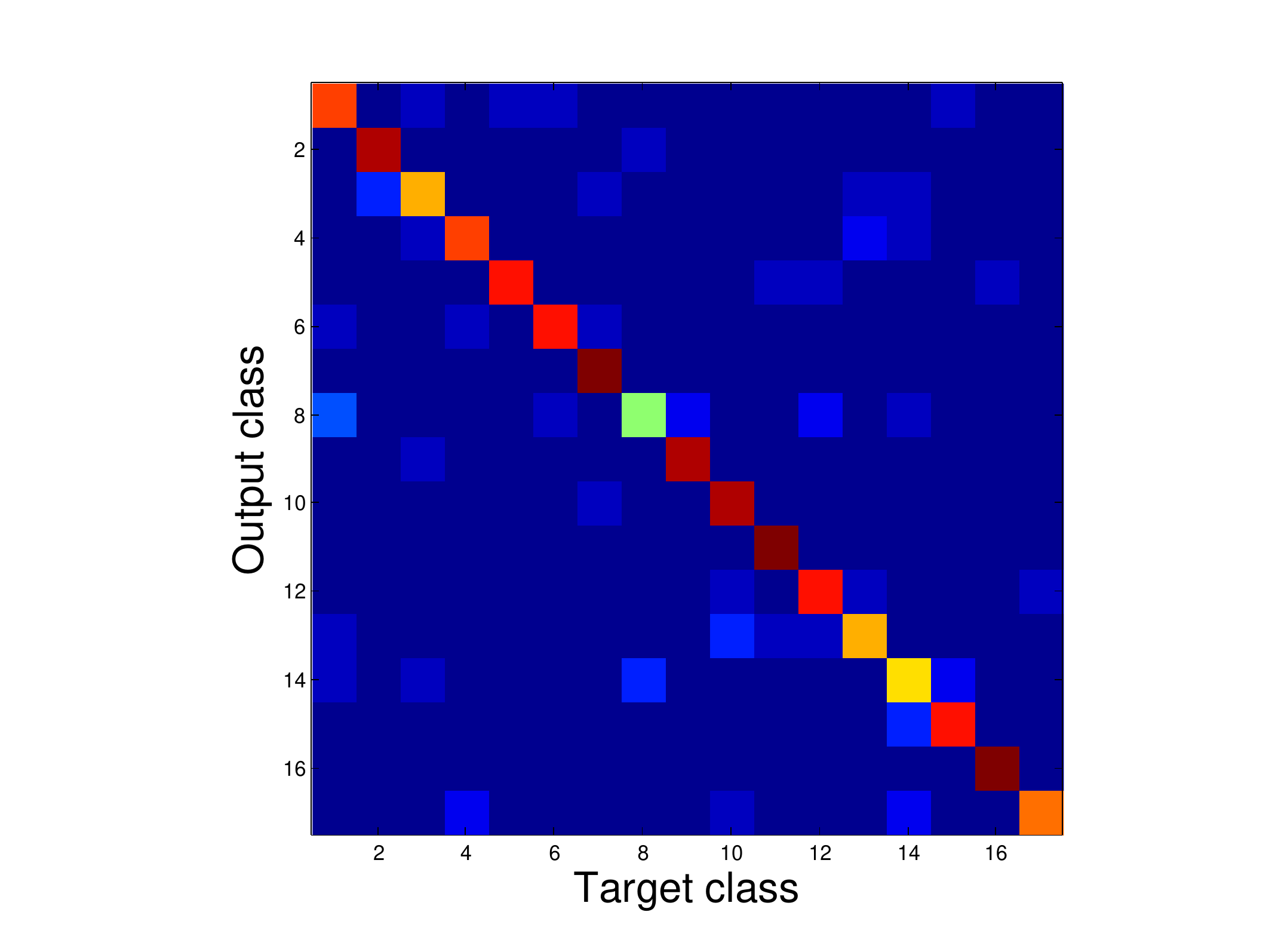}
\caption{ Confusion matrix of Oxford 17 flower classification using the proposed \textbf{USDH}.}
\label{fig4:}
\end{figure}

Different from supervised hashing method, \textbf{USDH} learns hashing codes without label information. Thus, it has more practical potential which benefits not only image retrieval but also other computer vision tasks such as fine-grained classification. To verify it, we conduct experiments on fine-grained classification compared with some traditional features, such as, SIFT, HOG, HSV and so on. It is worth mentioning that DeepBit is also a deep unsupervised hashing method. However, DeepBit method \cite{lin2016learning} only requires hashing codes invariant to rotation and not considers the within-class variance among different images.
Fine grained classification is a classic computer vision task and refers to discriminating categories of same sub-class belong to different super class. This task requires image descriptors invariant to within-class variance. More specially, for flower classification, within-class variances include color difference, shape deformation and pose. We select multi-svm as classifier and conduct experiments with different features. Table~\ref{tab4:} and Figure~\ref{fig4:} shows classification accuracy of these experiment shows the experiment results of \textbf{USDH}. Since within-class variance limits the efficiency of traditional color descriptor and hand-crafted shape descriptor, hashing codes learned by deep network has a superior performance, improved $10.7\%$ than SIFT-Internal feature. Compared with DeepBit, our method still improvs $6.2\%$. Additionally, our method is same fast as DeepBit method and more faster than traditional descriptor since it has low dimension. From the above experiment, the proposed \textbf{USDH} method has been proved efficiency for classification task.



\section{Conclusions}
In this paper, we propose a novel unsupervised deep hashing method, named unsupervised semantic deep hashing method. The parameters of deep neural network is fine-tuned according to four loss function: 1) semantic loss; 2) quantization loss; 3) information loss; and 4) rotation loss. Compare with previous unsupervised deep hashing methods, \textbf{USDH} requires hashing codes to preserve the relevant semantic information in the feature space. Extensive experiments on CIFAR-10 dataset and NUSWIDE dataset demonstrate that our proposed method outperforms existing unsupervised hashing method for image retrieval task. And the experimental results on Oxford17 dataset also prove that the hashing code learned by \textbf{USDH} is also effective on other computer vision tasks, such as fine-grained classification.
\begin{table}
  \centering
  \caption{The recognition accuracy for fine grained classification on Oxford17 dataset
  compared with different features}
  \label{tab4:}
  \begin{tabular}{llc}
    \hline
     Feature &Accuracy & Training time(sec)\\
     \hline
     Colour  & 60.9 $\pm$ 2.1\%& 3\\
     Texture  & 70.2 $\pm$ 1.3\%& 4\\
     HOG  & 63.7 $\pm$ 2.7\%& 3\\
     HSV  & 58.5 $\pm$ 4.5\%& 4\\
     SIFT-Boundary  & 59.4 $\pm$ 3.3\%& 4\\
     SIFT-Internal  & 70.6 $\pm$ 1.6\%& 4\\
     DeepBit & 75.1 $\pm$ 2.5\%& 0.07\\
     \hline
     {\bfseries USDH} & {\bfseries 81.3 $\pm$ 2.1\% }& {\bfseries 0.07}\\
    \hline
  \end{tabular}
\end{table}

\bibliographystyle{IEEEbib}
\bibliography{icme2018template}

\end{document}